\documentclass[11pt]{article}

\usepackage[T1]{fontenc}
\usepackage{fullpage}
\usepackage{microtype}
\usepackage{amsmath,amssymb,amsthm}
\usepackage{mathtools}
\usepackage{booktabs}
\usepackage{array}
\usepackage{xcolor}

\RequirePackage[most]{tcolorbox}
\RequirePackage{colortbl}

\definecolor{citecolor}{HTML}{0071BC}
\definecolor{linkcolor}{HTML}{ED1C24}
\definecolor{mydarkblue}{rgb}{0,0.08,0.45}

\usepackage{hyperref}
\hypersetup{colorlinks=true, linkcolor=mydarkblue, citecolor=mydarkblue,urlcolor=mydarkblue}
\usepackage[round,authoryear]{natbib}
\usepackage[noabbrev,nameinlink]{cleveref}

\DeclareMathOperator{\MClip}{MClip}
\DeclareMathOperator{\Pol}{Polar}

\newcommand{\argmin}{\mathop{\mathrm{argmin}}}

\newcommand{\sign}{\mathop{\mathrm{sign}}}

\newcommand{\diag}{{\rm diag}}

\newcommand{\mup}{\ensuremath{\mu\mathrm{P}}}
\newcommand{\rms}{\mathrm{R}}
\newcommand{\MuCon}{\mathrm{MuCon}}
\newcommand{\SpectralP}{\textsc{SpectralP}}

\title{\bfseries MuCon: Clipped Muon Updates for LLM Training}
\author{Albert Yi}
\date{May 8, 2026}

\begin{document}
\maketitle
\vspace{-2ex}

\begin{abstract}
Muon-style optimizers take a matrix-valued momentum or preconditioned update
\[
B=U\diag(\sigma_1,\dots,\sigma_r)V^\top
\]
and replace it with its canonical partial polar factor
\[
\Pol(B)=UV^\top .
\]
This maps every nonzero singular value to one. MuCon is the clipped-Muon variant studied here: it applies singular-value clipping to the same Muon matrix,
\[
D^{\MuCon}_\tau(B)
=
\MClip_\tau(B)
=
U\diag(\min\{\sigma_i,\tau\})V^\top,
\qquad
\tau>0.
\]
Thus, $\MClip_\tau$ denotes the mathematical clipping operator, while MuCon denotes the optimizer primitive that substitutes this clipped direction for Muon's polar direction. The Muon/MuCon scaling parameterization used in this work is called \SpectralP{}: it is the hidden-matrix scaling recipe under which polar Muon or clipped MuCon directions are applied. The map $\MClip_\tau$ is the Frobenius projection onto the spectral-norm ball of radius $\tau$: it leaves singular values at or below $\tau$ unchanged and modifies only the violating singular directions.
This paper asks when the MuCon clipping step can be approximated without a full dense SVD. We record two exact identities, a polar/absolute-value formula and a scalar-root formulation leading to a rational Newton filter for the clipped positive-semidefinite factor, and identify the numerical obstruction common to both: singular values near the threshold make sign decisions and rational solves ill-conditioned. Matrix-function methods are therefore useful only when paired with stable polar/square-root primitives or explicit regularization near the clipping boundary.
\end{abstract}

\section{Introduction}

Many optimizer designs modify matrix-valued updates through spectral transformations. A well-known example is Muon-style orthogonalization. If
\[
B_t=U\Sigma V^\top
\]
is a compact SVD of the matrix-valued momentum or preconditioned update passed to Muon's matrix step, the mathematical target in this report is the canonical partial polar factor
\[
D_t^{\mathrm{Muon}}=\Pol(B_t)=UV^\top .
\]
For rank-deficient matrices, $\Pol$ denotes this SVD-defined partial isometry, not an arbitrary orthogonal completion. In implementations it is often approximated by a Newton-Schulz iteration. The operation is aggressive: every nonzero singular value is replaced by one.

This report studies the more selective map
\[
\MClip_\tau(M)=U\diag(\min\{\sigma_i,\tau\})V^\top,
\qquad
M=U\diag(\sigma_i)V^\top,
\]
so that
\[
\sigma_i(\MClip_\tau(M))=\min\{\sigma_i(M),\tau\}.
\]
The default threshold is $\tau=1$, and we write $\MClip(M)=\MClip_1(M)$ when no threshold is shown. Clipping preserves all singular values at or below $\tau$ and modifies only the violating directions. When this map replaces Muon's polar step on the Muon matrix $B_t$, the resulting clipped-Muon update is
\[
D_t^{\MuCon}
=
\MClip_{\tau_t}(B_t),
\qquad
D_t^{\mathrm{Muon}}
=
\Pol(B_t).
\]
Throughout the paper, $\MClip_\tau$ denotes the mathematical clipping operator, while MuCon denotes the optimizer primitive that applies this operator inside the Muon update pipeline. Separately, \SpectralP{} denotes the scaling parameterization used for the Muon/MuCon hidden-matrix groups in this work; it is not a new clipping map.

The operator $\MClip_\tau$ is exactly the Frobenius projection onto the spectral-norm ball
\[
\mathcal B_2(\tau)=\{X\in\mathbb R^{m\times n}:\|X\|_2\le \tau\}:
\qquad
\MClip_\tau(M)=\argmin_{X\in\mathcal B_2(\tau)}\frac12\|X-M\|_F^2.
\]
The exact algorithm is simple: compute an SVD, clip the singular values, and reconstruct the matrix. Its dense cost,
\[
O(mn\min(m,n)),
\]
is usually too high for repeated use inside an optimizer; see \citet{golub2013matrix} for standard background on dense matrix decompositions.

The central numerical question is therefore:
\begin{quote}
\textit{Can the MuCon clipping step be approximated accurately enough for optimizer use while avoiding a full dense SVD?}
\end{quote}

A key structural identity already suggests the correct regime split. Let
\[
\mathcal I_>=\{i:\sigma_i(M)>\tau\},
\qquad
k_>=|\mathcal I_>|.
\]
Then
\[
\MClip_\tau(M)
=
M-U_>\diag\bigl((\sigma_i-\tau)_{i\in\mathcal I_>}\bigr)V_>^\top,
\]
where $U_>,V_>$ contain only the singular vectors whose singular values exceed $\tau$. Hence, clipping is a rank-$k_>$ correction to $M$. When $k_>$ is small, a partial SVD, Lanczos method, or randomized subspace method is the most selective baseline; when $k_>$ is large, global matrix-function iterations may be competitive.

\paragraph{Contributions.}
The report makes three technical points for \SpectralP{} MuCon. First, it separates the mathematical clipping map from the clipped-Muon optimizer primitive and records the projection and low-rank correction identities that any approximation should respect. Second, it derives a polar/absolute-value formulation of clipping and explains why threshold eigenvalues are numerically delicate. Third, it analyzes a rational Newton iteration for the clipped positive-semidefinite factor and clarifies that it is a spectral filter, not a standalone SVD-free algorithm.

\section{Background: \SpectralP{} and Width-Depth Scaled Training}

This project is motivated by hyperparameter transfer under simultaneous width and depth scaling. CompleteP studies joint width-depth transfer and non-lazy feature learning in deep Transformers \citep{dey2025completep}. Spectral \mup{} and related operator-norm viewpoints motivate controlling weights and updates in normalized spectral norm \citep{yang2023spectral,zheng2026spectral}. In this report, the resulting hidden-matrix scaling parameterization for Muon and MuCon is called \SpectralP{}. \SpectralP{} assigns hidden two-dimensional matrix groups to a spectral update class, while scalar, vector, embedding, and unembedding groups remain AdamW companion groups. Both viewpoints point to the same numerical need: matrix updates should have controlled spectra without requiring expensive decompositions at every optimizer step.

\subsection{Hyperparameter Transfer and Maximal-Update Parameterization}

A central practical problem in large-model training is hyperparameter transfer. Ideally, one tunes hyperparameters such as learning rate, initialization scale, and weight decay on a small model and transfers them to a much larger one. If the same base hyperparameters remain close to optimal after scaling, one obtains a ``tune small, train large'' strategy.

Maximal update parameterization, or \mup{}, was introduced to make this possible under width scaling \citep{yang2021tensorprogramsiv,yang2022tensorprogramsv}. In its simplest form, \mup{} aims to preserve nontrivial feature learning as the width $N$ grows. If $h_\ell(x)\in\mathbb R^{d_\ell}$ is the hidden representation at layer $\ell$, the invariant scale is coordinate-wise, or equivalently RMS-normalized:
\[
\|h_\ell(x)\|_{\rms,d_\ell}=\Theta(1),
\qquad
\|\Delta h_\ell(x)\|_{\rms,d_\ell}=\Theta(1),
\qquad
\|a\|_{\rms,d}:=\frac{\|a\|_2}{\sqrt d}.
\]
Thus, the parameterization should avoid both lazy dynamics,
\[
\|\Delta h_\ell(x)\|_{\rms,d_\ell}\to 0,
\]
and unstable dynamics,
\[
\|\Delta h_\ell(x)\|_{\rms,d_\ell}\to \infty.
\]

For width-only scaling, a typical hidden-matrix AdamW learning-rate rule takes the form
\[
\eta_{\mathrm{hidden}}=\eta_{\mathrm{base}}m_N^{-1},
\qquad
m_N=\frac{N}{N_{\mathrm{base}}}.
\]
The exact exponent depends on the optimizer and parameterization, but the principle is that matrix updates must be rescaled so that their induced feature movement remains order one.

Modern foundation models do not scale only in width; they also become deeper. A useful parameterization should therefore preserve transfer as
\[
N\to\infty,
\qquad
L\to\infty,
\]
where $L$ is the number of residual blocks.

\subsection{CompleteP: Residual Scaling for Deep Transformers}

CompleteP studies joint width-depth scaling for pre-LN decoder-only Transformer language models \citep{dey2025completep}. Its starting point is the residual recursion
\[
h_{\ell+1}=h_\ell+L^{-\alpha}F_\ell(h_\ell),
\qquad
\ell=1,\dots,L,
\]
where $F_\ell$ is a residual block, such as an attention or MLP block. The exponent $\alpha\in[1/2,1]$ controls the depth scaling of each residual branch.

The two most important cases are
\[
\alpha=\frac12
\qquad\text{and}\qquad
\alpha=1.
\]
The choice $\alpha=1/2$ is natural from initialization stability. If residual increments are roughly independent and have comparable size, then the accumulated residual variance scales as
\[
\sum_{\ell=1}^L L^{-2\alpha}=L^{1-2\alpha}.
\]
Avoiding variance explosion, therefore, requires
\[
\alpha\ge \frac12.
\]

CompleteP argues that initialization stability is not enough. It advocates the stronger scaling
\[
\alpha=1,
\]
so that
\[
h_{\ell+1}=h_\ell+L^{-1}F_\ell(h_\ell).
\]
In practical scaling experiments one writes
\[
m_N=\frac{N}{N_{\mathrm{base}}},
\qquad
m_L=\frac{L}{L_{\mathrm{base}}},
\]
and uses
\[
h_{\ell+1}=h_\ell+m_L^{-1}F_\ell(h_\ell).
\]

\subsubsection{AdamW scaling in CompleteP}

CompleteP is not only a residual multiplier. It also specifies how model and optimizer hyperparameters should scale with $m_N$ and $m_L$. For hidden matrix weights, the initialization variance follows the width-\mup{} rule
\[
\operatorname{Var}(W_{\mathrm{hidden}})=\sigma_{\mathrm{base}}^2m_N^{-1}.
\]
For AdamW hidden matrix updates, the learning-rate rule summarized in CompleteP is
\[
\eta_{\mathrm{hidden}}^{\mathrm{AdamW}}
=
\eta_{\mathrm{base}}m_N^{-1}m_L^{\alpha-1}.
\]
Thus, under CompleteP with $\alpha=1$,
\[
\eta_{\mathrm{hidden}}^{\mathrm{AdamW}}=\eta_{\mathrm{base}}m_N^{-1}.
\]
The hidden matrix learning rate scales with width, but not with depth.

LayerNorm and bias learning rates scale as
\[
\eta_{\mathrm{LN}}=\eta_{\mathrm{base}}m_L^{\alpha-1},
\qquad
\eta_{\mathrm{bias}}=\eta_{\mathrm{base}}m_L^{\alpha-1}.
\]
Therefore, under CompleteP,
\[
\eta_{\mathrm{LN}}=\eta_{\mathrm{bias}}=\eta_{\mathrm{base}}.
\]
CompleteP also scales hidden-weight decay and AdamW's numerical $\epsilon$ parameter by parameter group:
\[
\lambda_{\mathrm{hidden}}=\lambda_{\mathrm{base}}m_N.
\]
For the pure CompleteP AdamW parameterization, the AdamW $\epsilon$ scaling is parameter-group dependent. With a general residual exponent $\alpha$, hidden-block AdamW groups use
\[
\epsilon_{\mathrm{hidden/residual}}
=
\epsilon_{\mathrm{base}}m_N^{-1}m_L^{-\alpha}.
\]
This CompleteP hidden/residual group includes hidden matrix AdamW groups, hidden-block LayerNorm parameters, hidden-block biases, and other hidden vector parameters. Under CompleteP with $\alpha=1$, this becomes
\[
\epsilon_{\mathrm{hidden/residual}}
=
\epsilon_{\mathrm{base}}m_N^{-1}m_L^{-1}.
\]
CompleteP embedding/unembedding parameters and the final LayerNorm instead use
\[
\epsilon_{\mathrm{emb/unemb}}
=
\epsilon_{\text{final LN}}
=
\epsilon_{\mathrm{base}}m_N^{-1}.
\]
In \SpectralP{}, the same AdamW $\epsilon$ rules apply to AdamW companion groups: hidden-block LayerNorm, bias, and vector companion groups use the CompleteP hidden/residual $\epsilon$, while embedding/unembedding and final-LayerNorm companion groups use the CompleteP embedding/unembedding $\epsilon$. The \SpectralP{} Muon/MuCon hidden matrix groups themselves do not use AdamW $\epsilon$.

Biases and LayerNorm gains are usually assigned zero decoupled weight decay in LLM implementations. If they are decayed, their coefficient should be treated as a separate vector-parameter hyperparameter.

\subsubsection{Complete feature learning}

CompleteP also emphasizes \emph{complete feature learning}: a good width-depth parameterization should not merely keep activations stable, but should also prevent the network from becoming effectively linearized around initialization as $N,L\to\infty$.

Let $h(\theta)$ be a representation depending on parameters $\theta$, and let $\theta_0$ be the initialization. The linearization of $h$ at $\theta_0$ is
\[
h^{\mathrm{lin},\theta}(\theta,\theta_0)
=
h(\theta_0)
+
\left\langle \nabla_\theta h(\theta_0),\theta-\theta_0\right\rangle.
\]
A representation is lazy with respect to $\theta$ if its update becomes asymptotically indistinguishable from the update of this linearization:
\[
\frac{
\left|\Delta_\theta h-\Delta_\theta h^{\mathrm{lin},\theta}\right|
}{
\left|\Delta_\theta h^{\mathrm{lin},\theta}\right|
}
=o(1).
\]

The role of $\alpha$ can be seen in a two-layer residual block
\[
h_{\ell+1}=h_\ell+L^{-\alpha}W_\ell^{(2)}W_\ell^{(1)}h_\ell.
\]
Under maximal-update scaling, suppose
\[
\Delta W_\ell^{(i)}=\Theta(L^{\alpha-1}),
\qquad i=1,2.
\]
Then the first-order contribution to $\Delta h_{\ell+1}$ has size
\[
\Theta(L^{-1}),
\]
while the second-order contribution has size
\[
\Theta(L^{\alpha-2}).
\]
Their ratio is therefore
\[
\Theta(L^{\alpha-1}).
\]
If $\alpha<1$, this ratio vanishes; if $\alpha=1$, it remains order one. This is the basic mechanism behind the CompleteP choice $\alpha=1$.

\subsection{Relevance to \SpectralP{} MuCon}

The preceding scaling arguments motivate spectral control of matrix-valued updates. Singular value clipping provides an explicit projection-based control primitive:
\[
\MClip_\tau(M)=U\diag(\min\{\sigma_i,\tau\})V^\top,
\qquad
M=U\diag(\sigma_i)V^\top.
\]
It enforces
\[
\|\MClip_\tau(M)\|_2\le \tau
\]
and solves
\[
\MClip_\tau(M)=\argmin_{\|X\|_2\le \tau}\frac12\|X-M\|_F^2.
\]

\SpectralP{} MuCon applies this primitive to the same matrix-valued update that \SpectralP{} Muon would polarize. This differs from Muon-style orthogonalization,
\[
\Pol(M)=UV^\top,
\]
which sends every nonzero singular value to $1$. Clipping instead applies
\[
\sigma_i\mapsto \min(\sigma_i,\tau),
\]
so it preserves all singular directions with $\sigma_i\le\tau$ and modifies only the violating directions.

This distinction matters for optimizer design. If an update matrix has a few unstable high-gain directions but many useful moderate directions, full orthogonalization can distort the update unnecessarily. Singular-value clipping gives the closest feasible matrix under the spectral-norm constraint. The computational obstacle is that exact clipping requires an SVD. The numerical question is therefore:
\[
\text{Can the MuCon direction } \MClip_\tau(B_t) \text{ be approximated accurately and cheaply without a full SVD?}
\]

\subsection{Why iterative \SpectralP{} MuCon is a natural next step}

CompleteP motivates depth scaling through
\[
h_{\ell+1}=h_\ell+L^{-1}F_\ell(h_\ell),
\]
while spectral \mup{} motivates controlling matrix weights and updates in normalized operator norm. Both views suggest that optimizer design increasingly depends on matrix spectral geometry.

\SpectralP{} MuCon fits naturally into this picture. It controls the largest singular values without forcing the update to become an isometry:
\[
B_t\quad\mapsto\quad \MClip_\tau(B_t)
\]
rather than
\[
B_t\quad\mapsto\quad \Pol(B_t)=UV^\top .
\]
This makes it a candidate primitive for optimizers that want the stability benefits of spectral control while preserving more of the original update geometry.

The rest of the report isolates the numerical linear algebra problem from the full training problem. The next section states the scaling conventions used in this report.

\section{\SpectralP{} Scaling Recipes and Operator-Norm Bookkeeping}
\label{sec:scaling-recipes}

Write
\[
m_N=\frac{N}{N_{\mathrm{base}}},
\qquad
m_L=\frac{L}{L_{\mathrm{base}}}.
\]
For a vector $a\in\mathbb R^d$, define the dimension-normalized RMS norm
\[
\|a\|_{\rms,d}=\frac{\|a\|_2}{\sqrt d}.
\]
For $A\in\mathbb R^{m\times n}$, viewed as a map $\mathbb R^n\to\mathbb R^m$, define the induced RMS operator norm
\[
\|A\|_{\rms(m,n)}
=
\max_{v\ne0}\frac{\|Av\|_{\rms,m}}{\|v\|_{\rms,n}}
=
\sqrt{\frac{n}{m}}\,\|A\|_2.
\]
At fixed aspect ratio, this is equivalent to the spectral norm up to constants. For input or output matrices with one fixed dimension, the factor $\sqrt{n/m}$ is essential.

Table~\ref{tab:scaling-recipes} summarizes the recipe used in this report. Each entry is the multiplier applied to the corresponding base-model hyperparameter. The base model has $m_N=m_L=1$. The CompleteP column specializes to residual exponent $\alpha=1$ and reports the pure CompleteP AdamW scaling. The last column is the \SpectralP{} recipe for Muon/MuCon hidden matrix groups together with its AdamW companion groups. For a general residual multiplier $m_L^{-\alpha}$, replace the hidden/LN/bias AdamW learning-rate depth factor by $m_L^{\alpha-1}$ and the hidden-block AdamW $\epsilon$ depth factor by $m_L^{-\alpha}$. In both CompleteP and \SpectralP{}, hidden-block AdamW $\epsilon$ applies to hidden residual matrix, hidden LayerNorm, hidden bias, and hidden vector groups; embedding/unembedding and final LayerNorm groups keep the $m_N^{-1}$ width factor with no depth factor. In \SpectralP{}, hidden 2D Muon/MuCon matrix groups do not use AdamW $\epsilon$.

The implementation also follows the SteptronOSS Muon convention of applying a shape-dependent RMS-matching multiplier after the spectral direction is computed. For a hidden matrix with last two dimensions $m\times n$, define
\[
\kappa_{\mathrm{Muon}}(m,n)
=
\rho_{\mathrm{match}}\sqrt{\max\{m,n\}},
\]
where $\rho_{\mathrm{match}}$ is the configured \texttt{matched\_adamw\_rms}. Thus the \SpectralP{} optimizer group still has learning-rate multiplier one, but the actual hidden-matrix Muon step uses the effective coefficient $\eta_{\rm base}\kappa_{\mathrm{Muon}}(m,n)$ multiplying the polar or clipped direction. This is an implementation-level RMS matching calibration, not an AdamW $\epsilon$ rule and not a width/depth scheduler multiplier.

For embedding and unembedding AdamW companion groups, write $\gamma_{\rm emb}$ for the configured embedding learning-rate multiplier. The implementation therefore uses embedding/unembedding learning rate $\gamma_{\rm emb}\eta_{\rm base}$, with $\gamma_{\rm emb}=1$ as the default.

\begin{table}[ht!]
\centering
\scriptsize
\setlength{\tabcolsep}{3.0pt}
\renewcommand{\arraystretch}{1.18}
\begin{tabular}{
>{\raggedright\arraybackslash}p{0.24\linewidth}
>{\raggedright\arraybackslash}p{0.23\linewidth}
>{\raggedright\arraybackslash}p{0.25\linewidth}
>{\raggedright\arraybackslash}p{0.22\linewidth}}
\toprule
Quantity
&
\mup{} AdamW
&
CompleteP AdamW
&
\SpectralP{} Muon / MuCon
\\
\midrule

Scope
&
Width-only transfer; take $m_L=1$.
&
Joint width-depth AdamW transfer with residual exponent $\alpha=1$.
&
\SpectralP{} joint width-depth transfer for hidden 2D matrix groups using either a polar/Muon direction or the clipped-Muon direction. 
\\

Residual hidden-block multiplier
&
$1$
&
$m_L^{-1}$
&
$m_L^{-1}$
\\

Output/unembedding forward multiplier
&
$m_N^{-1}$
&
$m_N^{-1}$
&
$m_N^{-1}$
\\

Hidden matrix initialization variance
&
$\sigma_{\rm base}^2 m_N^{-1}$
&
$\sigma_{\rm base}^2 m_N^{-1}$
&
$\sigma_{\rm base}^2 m_N^{-1}$
\\

Input/embedding initialization variance
&
$\sigma_{\rm base}^2$ for one-hot input; $\sigma_{\rm base}^2/d_{\rm in}$ for dense input.
&
same
&
same
\\

Output/unembedding initialization variance
&
$\sigma_{\rm base}^2$
&
$\sigma_{\rm base}^2$
&
$\sigma_{\rm base}^2$
\\

Hidden matrix learning rate
&
$\eta_{\rm base}m_N^{-1}$
&
$\eta_{\rm base}m_N^{-1}$
&
$\eta_{\rm base}$ group multiplier; effective matrix step coefficient $\eta_{\rm base}\kappa_{\mathrm{Muon}}(m,n)$
\\

Embedding/unembedding learning rate
&
$\gamma_{\rm emb}\eta_{\rm base}$
&
$\gamma_{\rm emb}\eta_{\rm base}$
&
AdamW companion group: $\gamma_{\rm emb}\eta_{\rm base}$
\\

Bias/LayerNorm learning rate
&
$\eta_{\rm base}$
&
$\eta_{\rm base}$
&
AdamW companion group: $\eta_{\rm base}$
\\

Hidden matrix weight decay
&
$\lambda_{\rm base}m_N$
&
$\lambda_{\rm base}m_N$
&
$\lambda_{\rm base}$ for decoupled decay on \SpectralP{} Muon/MuCon hidden matrix groups
\\

Embedding/unembedding weight decay
&
$\lambda_{\rm base}$
&
$\lambda_{\rm base}$
&
AdamW companion group: $\lambda_{\rm base}$
\\

Bias/LayerNorm weight decay
&
Usually $0$ in decoupled-WD LLM implementations.
&
Usually $0$; if deliberately decayed, tune as a vector-parameter.
&
AdamW companion group.
\\

AdamW $\epsilon$, hidden residual matrix, hidden LayerNorm, and hidden bias/vector groups
&
$\epsilon_{\rm base}m_N^{-1}$
&
\textbf{CompleteP: } $\epsilon_{\rm base}m_N^{-1}m_L^{-1}$
&
\textbf{\SpectralP{}:} not used for Muon/MuCon hidden matrix groups; AdamW companion hidden LayerNorm, hidden bias, and hidden vector groups use $\epsilon_{\rm base}m_N^{-1}m_L^{-1}$
\\

AdamW $\epsilon$, embedding/unembedding and final LayerNorm groups
&
$\epsilon_{\rm base}m_N^{-1}$
&
\textbf{CompleteP: } $\epsilon_{\rm base}m_N^{-1}$
&
\textbf{\SpectralP{}:} AdamW companion embedding/unembedding and final LayerNorm groups use $\epsilon_{\rm base}m_N^{-1}$
\\

Matrix spectral post-processing
&
None.
&
None.
&
\SpectralP{} Muon uses the polar direction $D=\Pol(B)=UV^\top$ for hidden 2D matrices. \SpectralP{} MuCon uses the clipped-Muon direction $D=\MClip_\tau(B)$.
\\

\bottomrule
\end{tabular}
\caption{
Scaling recipe comparisons.
}
\label{tab:scaling-recipes}
\end{table}

\paragraph{Untied versus tied embeddings.}
Let $d_{\rm vocab}$ be fixed and store token matrices row-wise in hidden coordinates. In the untied setting, use separate matrices
\[
E,W_U\in\mathbb R^{d_{\rm vocab}\times N},
\qquad
h_0=E^\top e_x,
\qquad
\ell=m_N^{-1}W_Uh_L.
\]
The recipe is
\begin{align*}
\operatorname{Var}(E_{ij})=\operatorname{Var}((W_U)_{ij})=\sigma_{\rm base}^2,
\qquad
\eta_E=\eta_U=\gamma_{\rm emb}\eta_{\rm base},\\
\lambda_E=\lambda_U=\lambda_{\rm base},
\qquad
\epsilon_E=\epsilon_U=\epsilon_{\rm base}m_N^{-1}.
\end{align*}
In the tied setting, set $W_U=E$ and use one shared parameter and one shared optimizer state:
\begin{align*}
h_0=E^\top e_x,
\qquad
\ell=m_N^{-1}Eh_L,
\qquad
\operatorname{Var}(E_{ij})=\sigma_{\rm base}^2,
\qquad
\eta_E=\gamma_{\rm emb}\eta_{\rm base},\\
\lambda_E=\lambda_{\rm base},
\qquad
\epsilon_E=\epsilon_{\rm base}m_N^{-1}.
\end{align*}
The output multiplier remains $m_N^{-1}$ in both cases. In the tied case, this is especially important: the self-overlap term $E_x^\top E_x$ is order $N$ at initialization, so omitting this forward multiplier, or using a weaker width multiplier than $m_N^{-1}$, would make tied-output logits grow with width. Gradients from the input and output uses are summed before the single AdamW update. These embedding/unembedding groups are companion AdamW groups in the \SpectralP{} recipe, not hidden-matrix Muon/MuCon groups.

The operator-norm entries should be read as bookkeeping, not as a derivation of the AdamW column. In this report, \SpectralP{} names the Muon/MuCon hidden-matrix scaling rule summarized by the last column; MuCon itself remains the clipped update primitive. Let $G_{\mathcal O}(N)$ be the pre-learning-rate matrix direction produced by optimizer $\mathcal O$ before any implementation-level RMS matching multiplier. If, for a fixed aspect ratio,
\[
\|G_{\mathcal O}(N)\|_{\rms(m,n)}=\Theta(N^{p_{\mathcal O}}),
\]
then choosing
\[
\eta_{\mathrm{hidden}}^{\mathcal O}
=
\eta_{\mathrm{base}}m_N^{-p_{\mathcal O}}
\]
keeps the uncalibrated hidden matrix direction order one in the RMS operator norm. The exponent is update-normalization dependent and must be computed for the actual optimizer direction. For example, a dense i.i.d. $O(1)$ update has $\|G\|_2=\Theta(\sqrt N)$ and $p=1/2$ at fixed aspect ratio, whereas an uncalibrated polar/Muon hidden direction has $\|G\|_2=1$ and $p=0$. The code keeps this $p=0$ group-learning-rate convention, then multiplies the resulting Muon/MuCon direction by $\kappa_{\mathrm{Muon}}(m,n)$ to match the RMS scale used by the SteptronOSS Muon recipe. AdamW under \mup{}/CompleteP uses the coordinate-wise tensor-program scaling shown in Table~\ref{tab:scaling-recipes}; it should not be inferred by treating the AdamW update as a generic dense i.i.d. spectral update.

If threshold-$\tau$ clipping is applied to the pre-learning-rate hidden matrix direction, as in \SpectralP{} MuCon, then
\[
\|\MClip_\tau(G)\|_{\rms(m,n)}
\le
\sqrt{\frac{n}{m}}\,\tau.
\]
Thus, for hidden matrices with fixed aspect ratio and $\tau=\Theta(1)$, the uncalibrated \SpectralP{} MuCon direction is automatically order one in RMS operator norm. Under this bookkeeping, clipped MuCon has the same group-learning-rate width exponent as polar Muon, but its direction is not forced to be an isometry. The implemented step then applies the same $\kappa_{\mathrm{Muon}}(m,n)$ RMS-matching multiplier used for polar Muon.

\section{Problem Formulation}
\label{sec:problem-formulation}

Let
\[
M\in\mathbb R^{m\times n},
\qquad
M=U\Sigma V^\top
\]
be a compact SVD with rank $r$. Thus
\[
U\in\mathbb R^{m\times r},
\qquad
V\in\mathbb R^{n\times r},
\qquad
\Sigma=\diag(\sigma_1,\dots,\sigma_r),
\]
where
\[
\sigma_1\ge\sigma_2\ge\cdots\ge\sigma_r>0.
\]
The compact SVD omits zero singular values; all spectral formulas below act as zero on the omitted null spaces unless stated otherwise. For $M=0$, define $\MClip_\tau(0)=0$. For $M\neq0$, the singular-value clipping operator is
\[
\MClip_\tau(M)
=
U\diag(\min\{\sigma_1,\tau\},\dots,\min\{\sigma_r,\tau\})V^\top,
\qquad
\tau>0.
\]
Equivalently,
\[
\MClip_\tau(M)=Uf_\tau(\Sigma)V^\top,
\qquad
f_\tau(\sigma)=\min\{\sigma,\tau\}.
\]

In the optimizer setting, $M$ is the same matrix $B_t$ that \SpectralP{} Muon would pass to its polar or Newton-Schulz step. The \SpectralP{} MuCon update direction is therefore
\[
D_t^{\MuCon}
=
\MClip_{\tau_t}(B_t).
\]
This notation keeps the operator $\MClip_\tau$ distinct from the algorithmic primitive MuCon.

The projection interpretation is
\[
\MClip_\tau(M)
=
\argmin_{X\in\mathbb R^{m\times n}}
\frac12\|X-M\|_F^2
\quad
\text{subject to}
\quad
\|X\|_2\le \tau.
\]
Indeed, by unitary invariance of the Frobenius norm and von Neumann's trace inequality, the unique projection aligns its singular vectors with those of $M$ after zero-padding singular values as needed. The problem reduces to
\[
\min_{0\le s_i\le \tau}
\sum_i (s_i-\sigma_i)^2,
\]
whose unique solution is
\[
s_i=\min\{\sigma_i,\tau\}.
\]

The same formula gives the useful low-rank correction identity
\[
\MClip_\tau(M)
=
M-U_>\diag\bigl((\sigma_i-\tau)_{i\in\mathcal I_>}\bigr)V_>^\top,
\qquad
\mathcal I_>=\{i:\sigma_i>\tau\}.
\]
Consequently, if $k_>=|\mathcal I_>|\ll r$, the target differs from $M$ by a rank-$k_>$ matrix. This observation is central for algorithm design: global matrix-function approximations should be compared against partial SVD, Lanczos, or randomized range-finding baselines that target only the violating singular subspace \citep{halko2011finding}.

The computational challenge is that a full dense SVD is too expensive for frequent optimizer use. We therefore seek approximations using cheaper primitives:
\begin{itemize}
    \item matrix-matrix multiplications;
    \item matrix-vector products;
    \item matrix-function iterations;
    \item small dense auxiliary linear algebra; and
    \item structured linear solves when numerically stable.
\end{itemize}

\section{Algorithmic Approaches}

The low-rank correction identity suggests a partial spectral baseline whenever $k_>$ is small. For \SpectralP{} MuCon, this means targeting only the singular directions of the Muon matrix $B_t$ whose gains exceed $\tau_t$. The two approaches below are instead global matrix-function viewpoints. They are most relevant when many singular values violate the constraint, or when fast polar, square-root, or rational-filter primitives are already available.

\subsection{Approach I: polar/absolute-value formulation}

Let
\[
H=(M^\top M)^{1/2},
\qquad
Q=MH^\dagger .
\]
With the compact SVD above,
\[
Q=UV^\top,
\qquad
H=V\Sigma V^\top
\]
on the row space of $M$, and $H=0$ on its orthogonal complement. Thus
\[
M=QH.
\]
Functional calculus gives
\[
\MClip_\tau(M)=QP_\star,
\qquad
P_\star=f_\tau(H),
\qquad
f_\tau(t)=\min\{t,\tau\}.
\]
Since $f_\tau(0)=0$, this formula is valid for rectangular and rank-deficient matrices.

If $m<n$, it can be cheaper to work on the left side. With
\[
K=(MM^\top)^{1/2},
\qquad
Q=K^\dagger M=UV^\top,
\]
one has
\[
\MClip_\tau(M)=f_\tau(K)Q.
\]
Thus, the matrix-function factor should be formed on the smaller side whenever possible.

For a symmetric positive-semidefinite matrix $H$,
\[
P_\star
=
f_\tau(H)
=
\frac12\left(H+\tau I_n-|H-\tau I_n|\right),
\]
where $|A|=(A^2)^{1/2}$ for symmetric $A$. Here $f_\tau(H)$ denotes scalar functional calculus; it is not a Loewner infimum. Equivalently,
\[
|H-\tau I_n|
=
(H-\tau I_n)\sign(H-\tau I_n),
\]
with the spectral convention $\sign(0)=0$. Hence
\[
\MClip_\tau(M)=
Q\,\frac12\left(H+\tau I_n-(H-\tau I_n)\sign(H-\tau I_n)\right).
\]
This identity is exact, including at singular values equal to $\tau$.

This gives a two-stage strategy. First, approximate the canonical partial polar factor $Q$. For a tall full-column-rank matrix, the Newton-Schulz iteration
\[
X_{k+1}
=
\frac12X_k(3I-X_k^\top X_k)
\]
converges under standard scaling assumptions; a simple sufficient condition is $\|I-X_0^\top X_0\|_2<1$. Rank-deficient cases are more delicate in finite precision: exact zero singular values remain zero under the iteration, but tiny singular values can slow convergence and contaminate the computed partial polar factor. Rank-aware polar iterations, scaled Newton iterations, QDWH-type iterations, or explicit regularization are often preferable \citep{higham1986computing,nakatsukasa2010optimizing}. For wide matrices, apply the analogous iteration to $M^\top$ and transpose the result.

Second, form the symmetric factor
\[
\widehat H=\frac12(\widehat Q^\top M+M^\top\widehat Q),
\]
approximate $|\widehat H-\tau I_n|$, and set
\[
\widehat P
=
\frac12\left(\widehat H+\tau I_n-|\widehat H-\tau I_n|\right),
\qquad
\widehat X=\widehat Q\widehat P.
\]
For the left-sided variant, use
\[
\widehat K=\frac12(M\widehat Q^\top+\widehat Q M^\top),
\qquad
\widehat P_L=f_\tau(\widehat K),
\qquad
\widehat X=\widehat P_L\widehat Q.
\]
In finite precision, the symmetric factors should be explicitly symmetrized, and small negative eigenvalue artifacts should be treated as numerical error rather than meaningful spectrum.

\paragraph{Advantages.}
This formulation reduces clipping to classical matrix functions: polar decomposition and the matrix absolute value \citep{higham2008functions}.

\paragraph{Challenges.}
The clipped factor $f_\tau(H)$ is continuous but not differentiable at eigenvalue $\tau$. A standalone matrix-sign iteration for $H-\tau I_n$ is worse conditioned because the sign function is discontinuous at zero. The absolute-value formulation avoids this discontinuity, but it remains nonsmooth at the clipping boundary and can still be sensitive when many singular values satisfy $\sigma_i(M)\approx\tau$. Moreover, if $H$ is formed explicitly or if the absolute-value iteration requires dense eigensolves or large dense linear solves, the method can lose its advantage over SVD.

\subsection{Approach II: rational Newton iteration for the clipped PSD factor}

A second approach targets the clipped positive-semidefinite factor
\[
P_\star=f_\tau(H),
\qquad
H=(M^\top M)^{1/2}.
\]
For an eigenvalue $\sigma\ge0$ of $H$, the desired clipped value
\[
p_\star=\min\{\sigma,\tau\}
\]
is the smaller root, or the double root when $\sigma=\tau$, of
\[
(p-\tau)(p-\sigma)=0.
\]
Newton's method for this scalar equation gives
\[
p_{k+1}
=
\frac{p_k^2-\sigma\tau}{2p_k-\sigma-\tau},
\qquad
p_0=0.
\]
Let
\[
a=\min\{\sigma,\tau\},
\qquad
b=\max\{\sigma,\tau\},
\qquad
e_k=a-p_k.
\]
For $p_k<a$,
\[
e_{k+1}
=
\frac{e_k^2}{b-a+2e_k}.
\]
Thus the iterates increase to the smaller root. If $\sigma\ne\tau$, convergence is locally quadratic; if $\sigma=\tau$, then $e_{k+1}=e_k/2$ and convergence is only linear.

Applied by spectral functional calculus, this yields the rational matrix iteration
\[
(H+\tau I_n-2P_k)P_{k+1}=\tau H-P_k^2,
\qquad
P_0=0.
\]
Equivalently, in exact arithmetic,
\[
P_{k+1}
=
(H+\tau I_n-2P_k)^{-1}(\tau H-P_k^2),
\]
because $P_k=r_k(H)$ is a rational function of $H$ and therefore commutes with $H$. This formula should be interpreted as a rational spectral filter, not as a generic noncommutative Newton method. In finite precision, the linear solve should be implemented symmetrically and $P_{k+1}$ should be explicitly symmetrized.

Computing $P_\star$ alone does not recover the clipped matrix. One must also apply the polar factor:
\[
\MClip_\tau(M)=QP_\star=MH^\dagger P_\star,
\]
with the convention that the zero eigenspace of $H$ contributes zero in the pseudoinverse product. Equivalently,
\[
\MClip_\tau(M)=Mg_\tau(H),
\]
where
\[
g_\tau(\sigma)
=
\begin{cases}
1, & 0\le \sigma\le \tau,\\[1mm]
\tau/\sigma, & \sigma>\tau.
\end{cases}
\]
The value of $g_\tau$ at $\sigma=0$ is immaterial in exact arithmetic because $M$ annihilates the nullspace of $H$; the continuous convention $g_\tau(0)=1$ is preferable when approximating $g_\tau$ directly.

\paragraph{Advantages.}
The iteration targets the clipped positive-semidefinite factor directly, avoiding a separate discontinuous sign computation.

\paragraph{Challenges.}
The method still requires applying or approximating $H=(M^\top M)^{1/2}$ and then applying $H^\dagger P_\star$ or $Q$. It also involves solving 
\[
H+\tau I_n-2P_k.
\]
For the $i$th spectral component, the coefficient eigenvalue is
\[
\sigma_i+\tau-2p_{k,i}\longrightarrow |\sigma_i-\tau|.
\]
Thus, directions exactly at the clipping threshold make the limiting coefficient singular, and spectra clustered near $\sigma=\tau$ lead to ill-conditioned solves. The iteration is therefore not automatically cheaper than SVD unless combined with an efficient square-root/polar routine and stabilization near the threshold, such as smoothing the clip or regularizing the solve.

\section{Conclusion}

MuCon replaces Muon's polar direction with a clipped spectral direction. For the Muon matrix
\(B_t=U\Sigma V^\top\),
\[
D_t^{\mathrm{Muon}}=UV^\top,
\qquad
D_t^{\MuCon}=\MClip_{\tau_t}(B_t)
=
U\diag(\min\{\sigma_i,\tau_t\})V^\top.
\]
Thus MuCon is not a new parameterization: it is a projection primitive layered on top of the \SpectralP{} width-depth scaling and the implementation-level RMS-matching calibration. CompleteP supplies the residual and AdamW companion-group scaling rules; \SpectralP{} specifies the hidden-matrix Muon/MuCon group convention; clipping controls only the spectrum of the resulting matrix direction.

The main algorithmic lesson is a regime split. When only a few singular values exceed \(\tau\), clipping is the low-rank correction
\[
\MClip_\tau(M)
=
M-U_>\diag\bigl((\sigma_i-\tau)_{i\in\mathcal I_>}\bigr)V_>^\top,
\]
so partial SVD, Lanczos, or randomized subspace methods are the natural baselines. When many singular values are clipped, global matrix-function methods such as polar/absolute-value formulations become plausible, but only if the polar or square-root primitives are stable. Rational Newton filters are exact as spectral filters, but their solves become ill-conditioned near \(\sigma=\tau\), which explains their poor behavior without additional regularization.

% The 400-step nano arithmetic experiments are consistent with this view. The best tested exact-SVD MuCon setting, \(\texttt{mucon\_matched\_adamw\_rms}=8\) and \(\tau=2\), reaches validation loss \(0.0143\) and exact-sequence accuracy \(0.9734\), close to \SpectralP{} Muon accuracy \(0.9766\) but below the strongest AdamW baselines in this single-seed run. Subspace and polar/eigh MuCon track exact SVD closely, whereas rational Newton remains unstable as an optimizer backend. Overall, MuCon is best viewed as a stable spectral-control primitive whose practical value depends on matching the approximation method to the number and conditioning of the clipped singular directions.


\begin{thebibliography}{99}

\bibitem[Dey et~al.(2025)]{dey2025completep}
Nolan Dey, Bin Claire Zhang, Lorenzo Noci, Mufan Li, Blake Bordelon,
Shane Bergsma, Cengiz Pehlevan, Boris Hanin, and Joel Hestness.
\newblock Don't be lazy: CompleteP enables compute-efficient deep transformers.
\newblock \emph{arXiv preprint arXiv:2505.01618}, 2025.

\bibitem[Zheng et~al.(2026)]{zheng2026spectral}
Chenyu Zheng, Rongzhen Wang, Xinyu Zhang, and Chongxuan Li.
\newblock Spectral condition for $\mu$P under width-depth scaling.
\newblock \emph{arXiv preprint arXiv:2603.00541v1}, 2026.

\bibitem[Yang and Hu(2021)]{yang2021tensorprogramsiv}
Greg Yang and Edward J. Hu.
\newblock Tensor Programs IV: Feature learning in infinite-width neural networks.
\newblock In \emph{Proceedings of the International Conference on Machine Learning}, 2021.

\bibitem[Yang et~al.(2022)]{yang2022tensorprogramsv}
Greg Yang, Edward Hu, Igor Babuschkin, Szymon Sidor, Xiaodong Liu,
David Farhi, Nick Ryder, Jakub Pachocki, Weizhu Chen, and Jianfeng Gao.
\newblock Tensor Programs V: Tuning large neural networks via zero-shot hyperparameter transfer.
\newblock \emph{arXiv preprint arXiv:2203.03466}, 2022.

\bibitem[Yang et~al.(2023)]{yang2023spectral}
Greg Yang, James B. Simon, and Jeremy Bernstein.
\newblock A spectral condition for feature learning.
\newblock \emph{arXiv preprint arXiv:2310.17813}, 2023.

\bibitem[Jordan et~al.(2024)]{jordan2024muon}
Keller Jordan, Yuchen Jin, Vladimir Boza, Jiacheng You, Franz Cesista,
Laker Newhouse, and Jeremy Bernstein.
\newblock Muon: An optimizer for hidden layers in neural networks.
\newblock Technical blog post, 2024.

\bibitem[Loshchilov and Hutter(2019)]{loshchilov2019decoupled}
Ilya Loshchilov and Frank Hutter.
\newblock Decoupled weight decay regularization.
\newblock In \emph{International Conference on Learning Representations}, 2019.

\bibitem[Higham(1986)]{higham1986computing}
Nicholas J. Higham.
\newblock Computing the polar decomposition-with applications.
\newblock \emph{SIAM Journal on Scientific and Statistical Computing}, 7(4):1160-1174, 1986.

\bibitem[Higham(2008)]{higham2008functions}
Nicholas J. Higham.
\newblock \emph{Functions of Matrices: Theory and Computation}.
\newblock SIAM, 2008.

\bibitem[Nakatsukasa and Bai(2010)]{nakatsukasa2010optimizing}
Yuji Nakatsukasa and Zhaojun Bai.
\newblock Optimizing Halley's iteration for computing the matrix polar decomposition.
\newblock \emph{SIAM Journal on Matrix Analysis and Applications}, 31(5):2700-2720, 2010.

\bibitem[Golub and Van Loan(2013)]{golub2013matrix}
Gene H. Golub and Charles F. Van Loan.
\newblock \emph{Matrix Computations}.
\newblock Johns Hopkins University Press, 4th edition, 2013.

\bibitem[Halko et~al.(2011)]{halko2011finding}
Nathan Halko, Per-Gunnar Martinsson, and Joel A. Tropp.
\newblock Finding structure with randomness: Probabilistic algorithms for constructing approximate matrix decompositions.
\newblock \emph{SIAM Review}, 53(2):217-288, 2011.

\end{thebibliography}
\end{document}